\documentclass[lettersize,journal]{IEEEtran}
\usepackage{amsmath,amsfonts}
\usepackage{algorithmic}
\usepackage{algorithm}
\usepackage{array}
\usepackage[caption=false,font=normalsize,labelfont=sf,textfont=sf]{subfig}
\usepackage{textcomp}
\usepackage{stfloats}
\usepackage{url}
\usepackage{verbatim}
\usepackage{graphicx}
\usepackage{cite}

\usepackage{adjustbox}
\usepackage{booktabs}
\usepackage{tikz}
\usepackage{multirow}
\usepackage{orcidlink}

\usepackage[T1]{fontenc}
\usepackage{textcomp}

\begin{document}

\title{RaSFNet: Ray-aligned and Scan-aware Fusion Network for LiDAR Semantic Segmentation}

\author{Chuang Chen\orcidlink{0009-0008-0689-3985}
\thanks{($Corresponding \ author: Chuang\  Chen.$)}
\thanks{Chuang Chen is with the National Key Laboratory of Fundamental Science on Synthetic Vision, Sichuan University, Chengdu, 610065, China. (e-mail: chenchuang@stu.scu.edu.cn)}

\thanks{Code will be available on: \url{https://github.com/wtxjbl-tp/RaSFNet} }
}

\markboth{Journal of \LaTeX\ Class Files,~Vol.~14, No.~8, August~2021}%
{Shell \MakeLowercase{\textit{et al.}}: A Sample Article Using IEEEtran.cls for IEEE Journals}

\maketitle

\begin{abstract}
LiDAR point cloud semantic segmentation is a fundamental component of 3D perception for autonomous driving and robotics. Projection-based methods map 3D point clouds into 2D representations for feature extraction using 2D networks. However, aggregating multiple points during projection can obscure geometric differences within each frustum, while local feature extraction limits the modeling of spatial relationships across scanning regions. Moreover, proximity in the projected representation does not necessarily imply proximity in 3D space, and interactions between point and projected features may mix features from different ranges, reducing semantic discriminability. To address these challenges, we propose the Ray-aligned and Scan-aware Fusion Network (RaSFNet), which incorporates point cloud geometry and LiDAR scanning characteristics into projection-based semantic segmentation. Specifically, the Ray-Aligned Geometric Distribution Encoding (RAGDE) module characterizes point distributions within each frustum using a ray-aligned coordinate system. To model spatial context, a Scan-Topology-Aware Mamba (STAM) block organizes frustum features according to the LiDAR azimuth sampling pattern and captures long-range dependencies across scanning regions. In addition, a Range-Constrained Point-Frustum Fusion (RCPFF) mechanism uses shared range information to guide feature exchange between point and frustum representations while maintaining spatial correspondence. RaSFNet achieves mIoU scores of 70.2\% and 77.7\% on the SemanticKITTI and nuScenes validation sets, respectively, indicating competitive segmentation performance compared with the evaluated methods.
\end{abstract}


\section{Introduction}

\IEEEPARstart{L}{iDAR} point cloud semantic segmentation is a fundamental task for 3D scene understanding, providing fine-grained environmental perception for autonomous driving, robotic navigation, and intelligent transportation systems, and serving as an essential foundation for outdoor autonomous decision-making \cite{yan2025evaluating},\cite{ding2026clear}. However, LiDAR point clouds acquired in real-world outdoor environments are inherently characterized by irregular spatial distributions, sparsity, and highly non-uniform point densities, which make it challenging for existing models to effectively capture fine-grained geometric structures \cite{qi2017pointnet},\cite{hu2020randla},\cite{wang2024sfpnet}. Consequently, spatially adjacent ground classes with similar geometric characteristics, such as parking areas, sidewalks, and terrain, can be easily confused, limiting segmentation accuracy and the capability of existing models to meet the demands of complex outdoor perception \cite{abbasi20243d},\cite{kong2023rethinking},\cite{chen2024filling},\cite{xu2025frnet},\cite{FARVNet}. Therefore, developing feature modeling methods tailored to the inherent characteristics of LiDAR point clouds, which can jointly capture fine-grained geometric details and global spatial dependencies, is crucial for improving semantic segmentation performance in complex outdoor scenes and remains a key challenge in 3D LiDAR perception \cite{lai2023spherical},\cite{ma2026mfinet}.

Considering that LiDAR acquires three-dimensional structural information by continuously scanning the surrounding environment with multiple laser beams along the horizontal direction, the resulting point clouds preserve regular spatial indexing relationships in the angular domain \cite{behley2019semantickitti},\cite{Panoptic-nuScenes}. Building on this property, range-image methods \cite{milioto2019rangenet++},\cite{cortinhal2020salsanext},\cite{ando2023rangevit},\cite{kong2023rethinking} employ spherical projection to determine the row and column positions of each 3D point on a 2D grid according to its elevation and azimuth angles, transforming irregular point clouds into structured 2D representations. Such representations preserve the intrinsic scanning geometry and support the use of mature 2D backbone networks, providing structured inputs for LiDAR perception and making range-view methods a mainstream paradigm for point cloud semantic segmentation. However, projecting 3D point clouds onto a fixed-resolution 2D grid is inherently a lossy spatial compression process that cannot fully preserve the original three-dimensional structural information \cite{li2022prnet},\cite{xu2021rpvnet}. On the one hand, multiple points at different ranges may be mapped to the same pixel, causing projection conflicts and the loss of local geometric details. On the other hand, sparse LiDAR sampling can lead to insufficient feature information in critical regions. These issues introduce discrepancies between 2D pixel adjacency and the 3D spatial topology, making it difficult to accurately capture geometric relationships in complex scenes and limiting the model\textquotesingle s ability to effectively represent complete three-dimensional structures \cite{kong2023rethinking},\cite{chen2024filling},\cite{xu2025frnet},\cite{li2022prnet},\cite{xu2021rpvnet},\cite{chen2025farvnet}.

\begin{figure*}[!t]
    \centering
    \includegraphics[width=\linewidth]{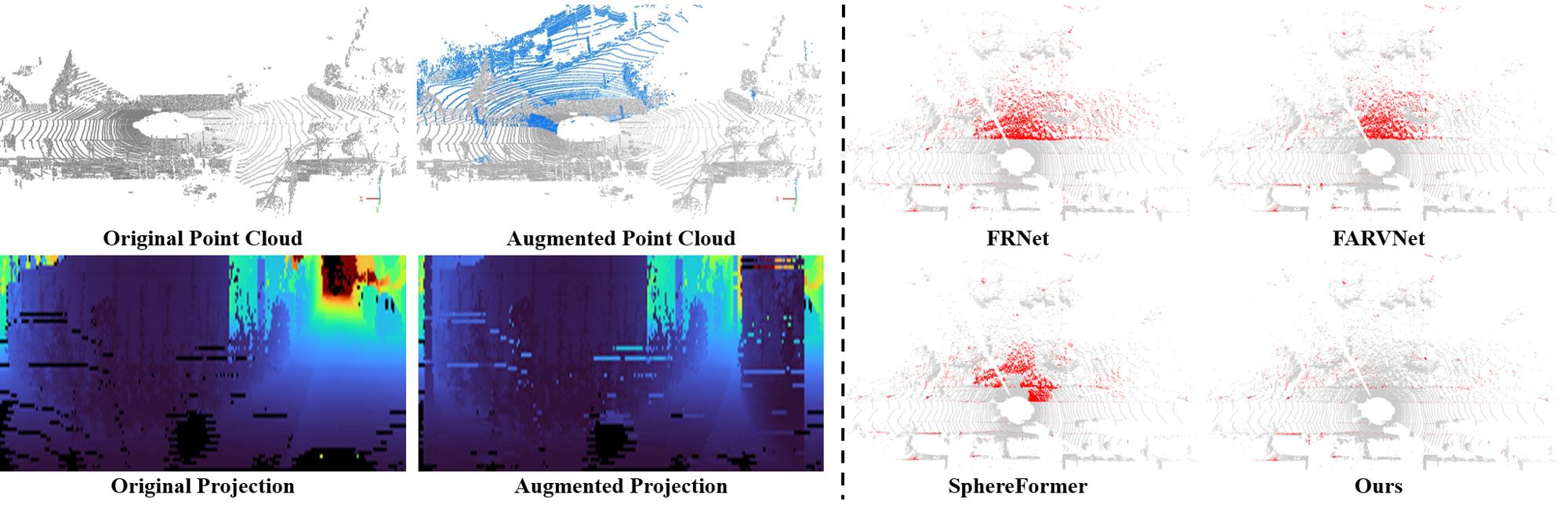}
    \caption{Visualization of projection conflicts caused by point cloud augmentation and their influence on semantic segmentation. The left part compares the original and augmented point clouds together with their corresponding range-view projections, while the right part shows the segmentation results of different methods in the affected regions. Red points indicate misclassified regions.}
    \label{existing_challenges}
\end{figure*}

Notably, the aforementioned projection issues are not limited to the representation of raw point clouds and may be further amplified by data augmentation strategies \cite{xu2025frnet}, \cite{ando2023rangevit}. A common strategy is to copy local point sets from other scenes into the current scene \cite{xiao2022polarmix}, \cite{chen2025copy}, enriching scene compositions and object distributions in the training data while alleviating overfitting to specific point cloud distributions. Although such methods can generate richer scene layouts in 3D space, they typically focus on the geometric plausibility of insertion positions while overlooking LiDAR ray visibility and occlusion constraints \cite{xu2025frnet}, \cite{chen2025farvnet}. In a LiDAR system, only the first valid return along a given laser direction is generally retained \cite{cortinhal2020salsanext},\cite{ando2023rangevit}. After external point sets are introduced, newly inserted points may share the same scanning direction as foreground points in the original scene. As illustrated in Fig. \ref{existing_challenges}, when the augmented point cloud is projected onto the range image, points with different ranges and semantic labels may fall into the same image cell, disturbing the local distributions of range, intensity, and geometric features.
To alleviate the information loss caused by projection conflicts, existing studies organize multiple points mapped to the same grid location into local frustum units and encode them as 2D frustum features \cite{xu2025frnet,chen2025farvnet,ma2026mfinet}. A bidirectional fusion mechanism is established between 3D point features and 2D frustum features, allowing the original point information in overlapping projection regions to continuously participate in iterative feature updates and alleviating the information compression caused by single-pixel representations. Despite this improvement, the intrinsic characteristics of LiDAR point clouds remain insufficiently exploited in 3D structure encoding, frustum feature extraction, and feature fusion, limiting the modeling of geometric and topological relationships in complex scenes and the discrimination of spatially adjacent structures with similar geometry, as illustrated in Fig. \ref{existing_challenges}.

Motivated by the above limitations, we propose a new method for collaborative learning between 3D structural features and 2D frustum features, termed the Ray-aligned and Scan-aware Fusion Network (RaSFNet), which enhances fine-grained 3D structure discrimination and improves the representation of geometrically similar objects in complex scenes. To distinguish different range distributions and geometric patterns within each frustum, we design a Ray-Aligned Geometric Distribution Encoding (RAGDE) module. Points are explicitly represented by their radial range and tangential offsets in a ray-aligned coordinate system, while multi-frequency sine and cosine functions are used for Fourier feature \cite{tancik2020fourier} mapping to enhance the representation differences among distinct spatial organization patterns. To enhance the global modeling capability of frustum features, we design a Scan-Topology-Aware Mamba (STAM) block. The module sequentially organizes features according to the LiDAR azimuth sampling pattern and leverages Mamba \cite{gu2024mamba} to capture long-range dependencies across scanning regions, improving the representation of structural variations among different scan regions and global spatial relationships. To address the decoupling between 3D structural features and 2D frustum features during deep interaction, we introduce a a Range-Constrained Point-Frustum Fusion (RCPFF) mechanism that uses distance information as a shared cue linking 2D projections and 3D spatial structures, guiding the collaborative update of both feature types while enhancing spatial consistency and semantic complementarity.

Overall, our main contributions are summarized as follows:

\begin{enumerate}
\item We propose RaSFNet, a network for collaborative learning between 3D structural features and 2D frustum features. Unlike existing point-frustum interaction methods that lack persistent spatial constraints, RaSFNet explicitly models the geometric structures within frustums and LiDAR scanning relationships, while using range information to constrain the interaction between point and frustum features.

\item We propose RAGDE, which uses each frustum ray as a reference and encodes radial range, tangential offsets, and local variations with multi-frequency Fourier features to distinguish geometric distributions.

\item We propose a STAM Block, which reorganizes frustum features into sequences according to the LiDAR azimuth scanning pattern, extending feature extraction from fixed local neighborhoods to global scan-wise dependencies.

\item We propose RCPFF, which uses distance information to constrain the interaction between 3D structural and 2D frustum features, guiding their complementary updates while maintaining spatial correspondence.

\item Comprehensive experiments on the SemanticKITTI \cite{behley2019semantickitti} and nuScenes \cite{Panoptic-nuScenes} validation sets show that RaSFNet achieves mIoU scores of 70.2\% and 77.7\%, respectively.

\end{enumerate}

\section{Related Work}
\subsection{Point Cloud Feature Encoding}
\label{Point Cloud Feature Encoding}

Point cloud feature encoding aims to extract discriminative 3D geometric information from the spatial distribution of raw point clouds. PointNet \cite{qi2017pointnet} uses shared multilayer perceptrons (MLPs) for point-wise encoding, while PointNet++ \cite{qi2017pointnet++} adds hierarchical local feature aggregation. RandLA-Net \cite{hu2020randla} strengthens neighborhood features through attentive local encoding. Cylinder3D \cite{zhu2021cylindrical} instead organizes points with cylindrical voxelization. Recent methods further improve point cloud representation from different perspectives. SFPNet \cite{wang2024sfpnet} employs sparse focal point modulation to aggregate multi-level contextual information, PointGeo \cite{an2025pointgeo} jointly captures local and global geometric features, and MDSLCA \cite{guang2026mdslca} combines multi-scale local attention with dilated spatial attention to enhance structural representations over different spatial ranges. Unlike methods that directly process unordered point sets, RangeNet++ \cite{milioto2019rangenet++} projects range, 3D coordinates, and remission values onto a regular 2D grid, while SphereFormer \cite{lai2023spherical} introduces radial windows to accommodate the distance-dependent sparsity of LiDAR point clouds. Although these methods enhance geometric feature encoding, their representations remain largely focused on local neighborhood structures, with insufficient explicit modeling of differences in range distributions, geometric patterns, and spatial organizations. Consequently, point sets with distinct structural characteristics may still produce similar encoded representations. In this work, we introduce RAGDE, which characterizes radial and tangential structures along the LiDAR sensing ray, while multi-frequency Fourier mapping improves the separability of local point distributions.

\subsection{Projected Image Feature Extraction}
\label{Projected Image Feature Extraction}

Projection-image feature extraction transforms point clouds into structured image representations and leverages 2D networks to learn semantic features at different scales. RangeViT \cite{ando2023rangevit} divides the projected image into feature patches and employs Transformers \cite{dosovitskiy2021image} to integrate local details with global semantics, while RangeFormer \cite{kong2023rethinking} adopts a hierarchical Transformer architecture to strengthen feature interactions across different regions. Fast FMVNet \cite{chen2024filling} improves range-image representation through an efficient multi-scale architecture, whereas MINet \cite{li2021multi} employs multiple feature paths at different scales with dense cross-scale interactions. RangeRet \cite{mosco2026revisiting} introduces circular retention to exploit the $360^\circ$ continuity of range images and model spatial dependencies through distance-aware decay. Nevertheless, the projected features in these methods are still organized primarily on regular 2D grids, with limited consideration of range and point-density variations across different azimuthal regions. In this work, we propose STAM, which reorganizes feature sequences according to the LiDAR azimuth sampling pattern and performs differentiated modeling across scanning regions.

\subsection{Point Cloud-Projected Image Feature Fusion}
\label{Point Cloud–Projected Image Feature Fusion}

To compensate for the information loss caused by a single projected representation, point-projection fusion methods retain the original point features and exploit the spatial correspondence between points and image representations for cross-representation interaction. PRNet \cite{li2022prnet} introduces range-view projection and point-range convolution to dynamically map range-image features back to the original points. FRNet \cite{xu2025frnet} employs point-frustum mappings to alternately propagate frustum features and aggregate point features across multiple network stages. HARP-NeXt \cite{abou2025harp} further develops a multi-scale point–range fusion backbone to enable interaction between 3D geometric information and 2D semantic information at different feature levels. MFINet \cite{ma2026mfinet} constructs 3D point-view, bird\textquotesingle s-eye-view, and range-view branches, enabling bidirectional 3D-2D interaction through point-feature projection and 2D feature enhancement while integrating complementary information across different views. However, these methods mainly rely on fixed point–projection correspondences for feature fusion, without maintaining a shared geometric constraint during feature updates. In this work, we propose a RCPFF mechanism that treats the range information shared by both representations as a common geometric anchor and explicitly incorporates it into feature interaction and updating, preserving spatial correspondence while enhancing the complementary representation of 3D geometric and 2D semantic information.

\section{Methodology}

\begin{figure*}[!t]
    \centering
    \includegraphics[width=\linewidth]{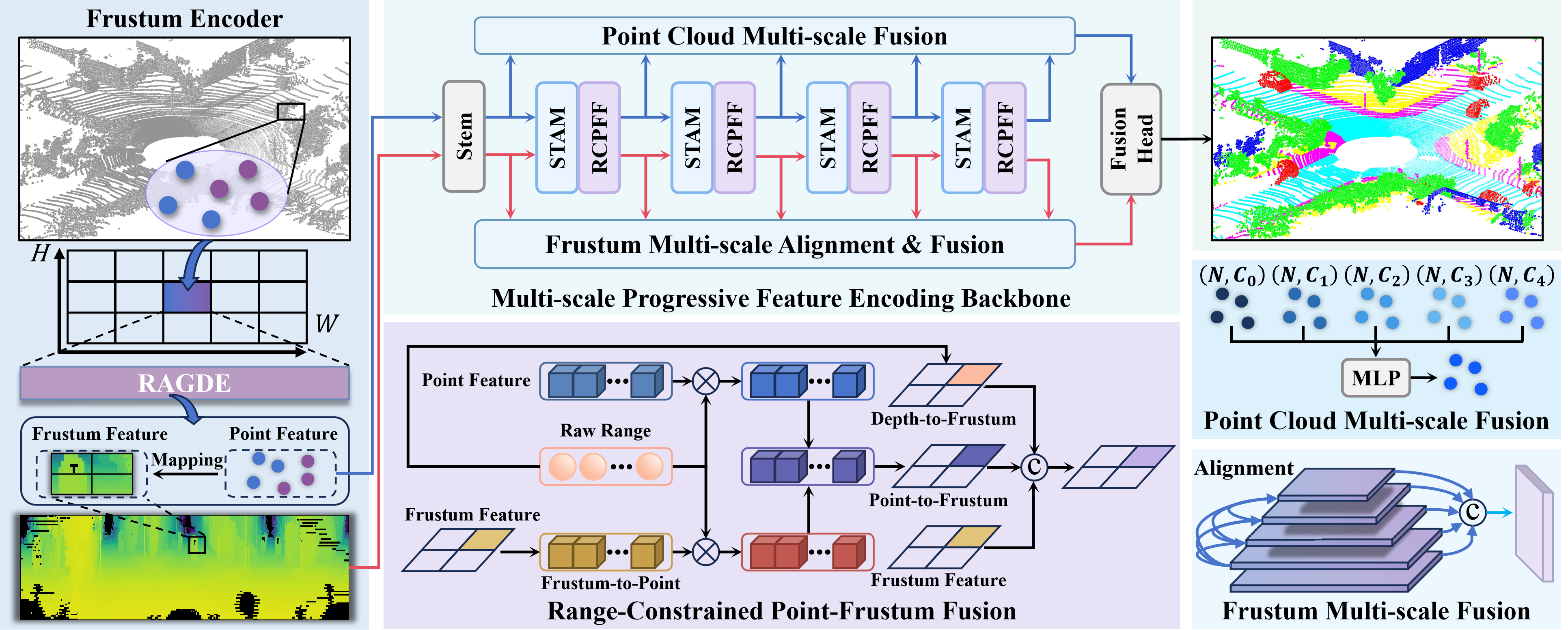}
    \caption{Overall architecture of the proposed RaSFNet. The network consists of RAGDE, STAM, RCPFF, and a prediction head. The input point cloud is encoded into 3D structural features and 2D frustum features, which are progressively updated through hierarchical feature extraction and cross-representation interaction, followed by feature fusion and final semantic prediction.}
    \label{flowchart}
\end{figure*}

RaSFNet enables collaborative learning between 3D structural representations and 2D frustum representations through multi-stage feature interaction, as illustrated in Fig. \ref{flowchart}. Given an input point cloud, spatial correspondences between points and frustums are first established according to the projection relationship, producing the corresponding 3D and 2D representations. RAGDE, STAM, and RCPFF are then integrated into successive feature stages to encode, update, and fuse the two types of features. Finally, the fused features are mapped back to the point space through the prediction head to generate point-wise semantic labels.

\subsection{Problem Definition}

Given an input LiDAR point cloud $\mathcal{P}=\{(\mathbf{p}_i,r_i)\}_{i=1}^{N}$, where $N$ denotes the number of points, $\mathbf{p}_i=[x_i,y_i,z_i]^{\mathrm{T}}\in\mathbb{R}^{3}$ represents the 3D coordinate of the $i$-th point, and $r_i$ denotes the corresponding remission intensity. Each point is mapped onto a 2D grid through spherical projection:
\begin{equation}
    (u_i,v_i)=\Pi(\mathbf{p}_i)
\end{equation}

where $\Pi(\cdot)$ denotes the LiDAR spherical projection function and $(u_i,v_i)$ represents the corresponding 2D grid coordinate. According to the projection coordinates, points mapped to the same grid cell are grouped into a frustum unit. Each frustum unit is associated with a 2D grid location and retains the indices of its constituent 3D points. The corresponding point and frustum features are represented as:
\begin{equation}
    \mathbf{F}^{3D}\in\mathbb{R}^{N\times C_{3D}}
    \qquad
    \mathbf{F}^{2D}\in\mathbb{R}^{H\times W\times C_{2D}}
\end{equation}

where $\mathbf{F}^{3D}$ denotes the 3D structural features associated with the original points, while $\mathbf{F}^{2D}$ denotes the 2D frustum features organized on an $H\times W$ projected grid. 

\subsection{Ray-Aligned Geometric Distribution Encoding}
\label{Ray-Aligned Geometric Distribution Encoding}

\begin{figure}[!h]
    \centering
    \includegraphics[width=\linewidth]{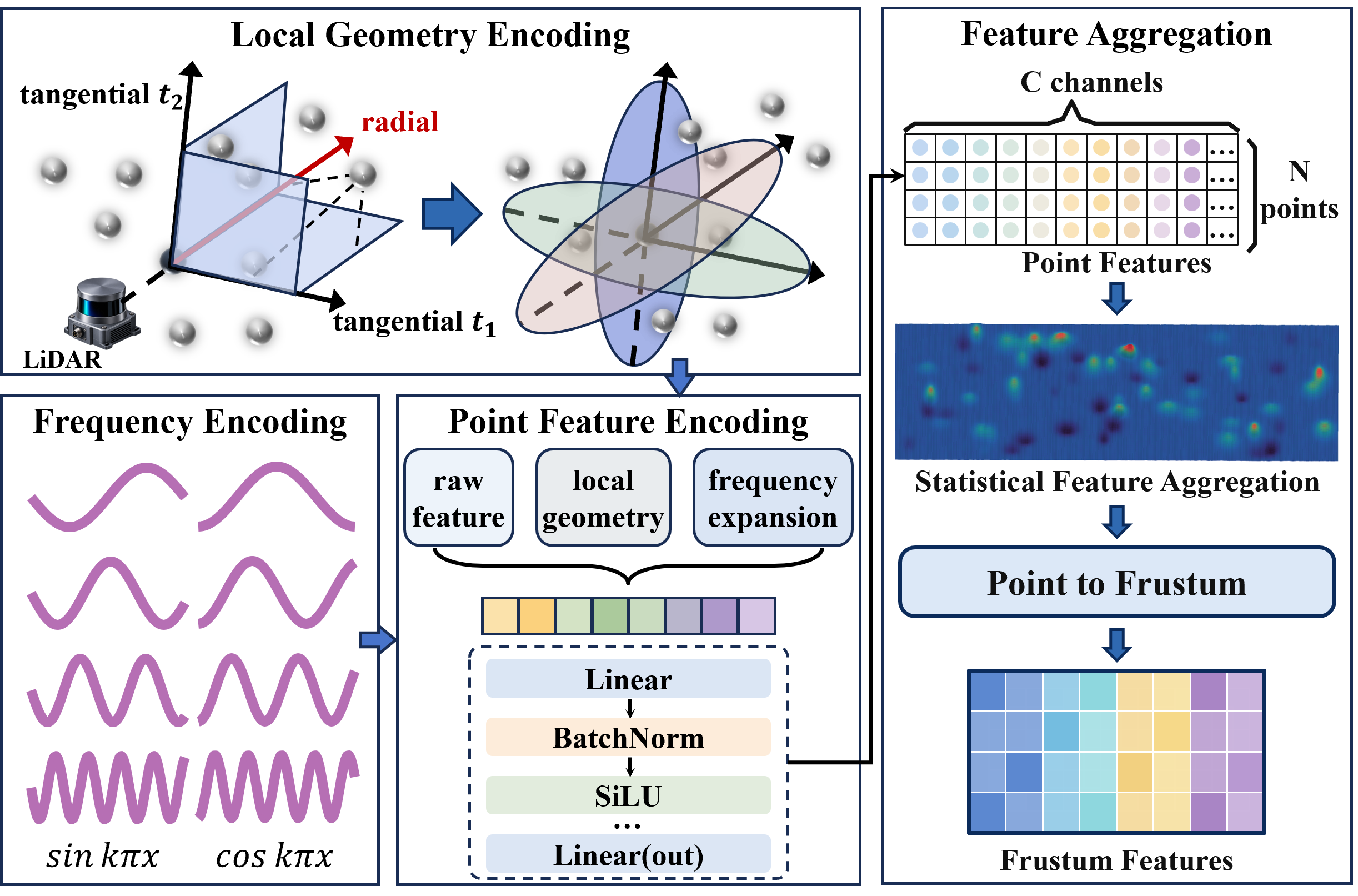}
    \caption{Architecture of the RAGDE module for encoding ray-aligned point distributions.
}
    \label{RAGDE}
\end{figure}

Points projected to the same image cell may exhibit different range distributions and geometric structures, while conventional pooling operations tend to suppress such structural differences \cite{xu2025frnet}. To address this issue, we propose the RAGDE module, as shown in Fig.~\ref{RAGDE}, to enhance the discrimination of intra-frustum 3D structures.

For the $g$-th frustum point set $\mathcal{P}_g=\{(\mathbf{p}_i,r_i)\}_{i=1}^{N_g}$, we first compute its spatial center $\bar{\mathbf{p}}_g$, with the direction from the LiDAR origin to this center serving as the radial axis $\mathbf{e}_r^g$. Two orthogonal tangential axes, $\mathbf{e}_{t_1}^g$ and $\mathbf{e}_{t_2}^g$, are further constructed on the plane perpendicular to the radial direction, under which the ray-aligned local coordinate of point $\mathbf{p}_i$ is expressed as:
\begin{equation}
    \mathbf{q}_i=
    \left[
    (\mathbf{p}_i-\bar{\mathbf{p}}_g)^{\mathrm T}\mathbf{e}_r^g,
    (\mathbf{p}_i-\bar{\mathbf{p}}_g)^{\mathrm T}\mathbf{e}_{t_1}^g,
    (\mathbf{p}_i-\bar{\mathbf{p}}_g)^{\mathrm T}\mathbf{e}_{t_2}^g
    \right]^{\mathrm T}
\end{equation}

To amplify subtle differences among similar local geometric features, logarithmic standard deviations $\boldsymbol{\sigma}_q^g$ are used to preserve directional scale information, while frustum-wise standardization and fixed-range clipping convert $\mathbf{q}_i$ into $\tilde{\mathbf{q}}_i$ for multi-frequency encoding:
\begin{equation}
\mathbf{q}_i^{\mathrm{mf}}
=
\operatorname{Concat}_{b=0}^{B-1}
\left[
\sin(2^b\pi\tilde{\mathbf q}_i),
\cos(2^b\pi\tilde{\mathbf q}_i)
\right], B=4
\end{equation}

The point attributes are first augmented with range information and intra-frustum coordinate and intensity offsets. These augmented attributes are concatenated with the $\tilde{\mathbf q}_i$, $\log(1+\boldsymbol{\sigma}_q^g)$, and $\mathbf{q}_i^{\mathrm{mf}}$ processed by point-wise feature layers to obtain $\mathbf F^{3D}$. 

Given the point-level features $\mathbf{F}^{3D}$, RAGDE constructs an initial frustum representation through distribution-aware aggregation. We compute the channel-wise mean and standard deviation to preserve both the overall feature response and its variability. To incorporate explicit spatial information, local coordinate statistics capture point locations and directional spread. These statistics are combined with range distribution, point count, and intensity statistics to form an auxiliary descriptor $\mathbf{s}_{(u_i,v_i)}$. Points sharing the same projected coordinates $(u_i,v_i)$ are aggregated into a single frustum feature, which is assigned to the corresponding grid location:
\begin{equation}
\mathbf{F}^{2D}[:,v_i,u_i]
=
\phi_f\left(
\left[
\boldsymbol{\mu}_{(u_i,v_i)},\,
\boldsymbol{\sigma}_{(u_i,v_i)},\,
\mathbf{s}_{(u_i,v_i)}
\right]
\right),
\end{equation}

where $\boldsymbol{\mu}_{(u_i,v_i)}$ and
$\boldsymbol{\sigma}_{(u_i,v_i)}$ denote the channel-wise mean and standard deviation of the point-level features projected to $(u_i,v_i)$, respectively. The operator $[\cdot]$ denotes concatenation, and $\phi_f$ consists of a linear layer, BatchNorm, and SiLU activation.

\subsection{Scan-Topology-Aware Mamba}
\label{sec:STAM}

\begin{figure}[!h]
    \centering
    \includegraphics[width=\linewidth]{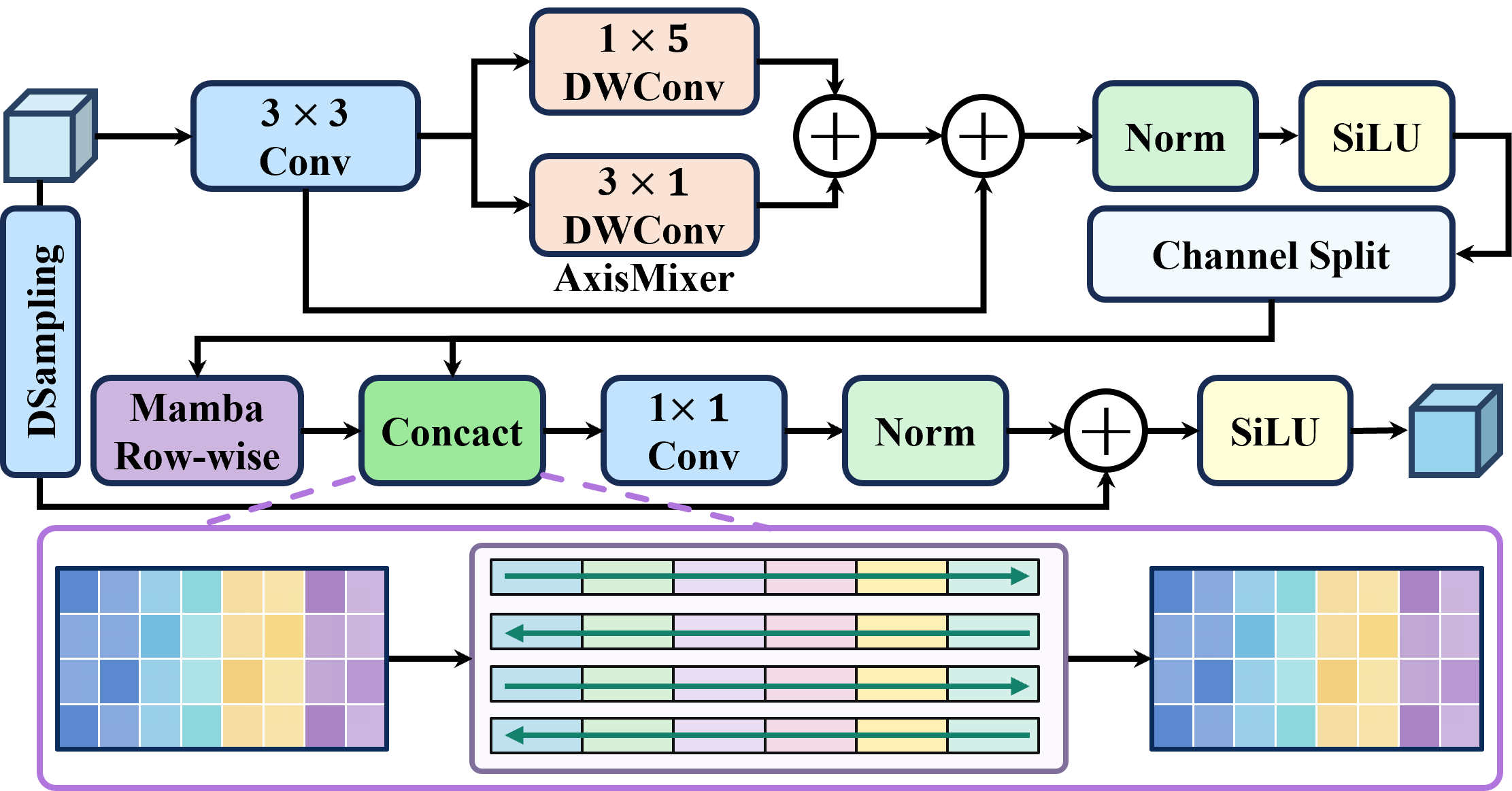}
    \caption{Architecture of STAM for anisotropic feature modeling.
}
    \label{STAM}
\end{figure}

The horizontal axis of $\mathbf{F}^{2D}$ follows the $360^\circ$ LiDAR azimuth scan, with its first and last columns corresponding to adjacent directions, whereas the vertical axis indexes different laser beams. To accommodate this anisotropic structure, STAM combines axis-aware local feature aggregation with sequence modeling over a subset of channels to capture feature dependencies along both axes, as shown in Fig.~\ref{STAM}.

STAM maps the input frustum feature $\mathbf{F}^{2D}$ into the target feature space through a $3\times3$ convolution. To account for scale differences between the horizontal azimuth and vertical laser-beam axes, axis-specific local aggregation is performed using range-wise convolutions \cite{scribano2026accelerating} with kernel sizes of $1\times5$ and $3\times1$, respectively:
\begin{equation}
\begin{aligned}
    \mathcal{A}(X)
    =
    \alpha_h\mathcal{C}_{1\times5}(X)
    +
    \alpha_v\mathcal{C}_{3\times1}(X),
    \\
    [\alpha_h,\alpha_v]
    =
    \operatorname{Softmax}([a_h,a_v]),\\
     \mathbf{F}_{Axis}
    =\mathcal{A}(\mathcal{C}_{3\times3}(\mathbf{F}^{2D}))
   , \\
    \mathbf{F}_l
    =
    \delta\left(
    \operatorname{Norm}
    \left(
    \mathbf{F}^{2D}+\mathbf{F}_{Axis}
    \right)
    \right)
\end{aligned}
\end{equation}

where the learnable parameters $a_h$ and $a_v$ balance the horizontal and vertical aggregation branches. STAM partitions the local feature $\mathbf{F}_l$ along the channel dimension into a row-wise Mamba branch $\mathbf{F}_m$ and a retained branch $\mathbf{F}_{\mathrm{retain}}$:
\begin{equation}
    [\mathbf{F}_m,\mathbf{F}_{retain}]
    =
    \operatorname{Split}_c(\mathbf{F}_l).
\end{equation}

$\mathcal{A}(\cdot)$ injects neighborhood context into $\mathbf{F}_m$, while $\mathcal{S}(\cdot)$ arranges the feature map into serpentine row-wise sequences and reverses the traversal direction between adjacent rows. STAM routes $\mathbf{F}_{retain}$ through a bypass connection to preserve local features without sequence modeling. The Mamba branch updates $\mathbf{F}_m$ as:
\begin{equation}
\begin{aligned}
    \mathbf{F}_{m}^{\prime}
    =
    \mathbf{F}_{m}
    +
    \gamma\,\mathcal{S}^{-1}
    \left(
    \operatorname{Mamba}
    \left(
    \operatorname{LN}
    \left(
    \mathcal{S}
    \left(
    \mathcal{A}(\mathbf{F}_{m})+\mathbf{F}_{m}
    \right)
    \right)
    \right)
    \right),
\end{aligned}
\end{equation}

where $\mathcal{S}(\cdot)$ and $\mathcal{S}^{-1}(\cdot)$ denote scan-aware serialization and spatial restoration, respectively, while the learnable scalar $\gamma$ controls the Mamba update. STAM concatenates $\mathbf{F}_{m}^{\prime}$ with $\mathbf{F}_{retain}$, applies a $1\times1$ convolution for channel fusion, and adds the residual feature $\mathcal{R}(\mathbf{F}^{2D})$ to produce the output:
\begin{equation}
\mathbf{F}_{stam}
=
\delta\left(
\operatorname{Norm}
\left(
\mathcal{C}_{1\times1}
\left(
[\mathbf{F}_{m}^{\prime},\mathbf{F}_{retain}]
\right)
\right)
+
\mathcal{R}(\mathbf{F}^{2D})
\right),
\end{equation}

where $\mathcal{R}(\cdot)$ performs downsampling along the residual branch.

\subsection{Range-Constrained Point-Frustum Fusion}
\label{sec:RCPFF}

RCPFF uses range information as a shared geometric reference to constrain bidirectional updates between $\mathbf{F}^{3D}$ and $\mathbf{F}_{stam}$. Given the frustum feature $\mathbf{F}_{stam}$, the point-level feature $\mathbf{F}^{3D}$, and the point-wise range $range$, the Frustum-to-Point operation retrieves the frustum feature at $(u,v)$ and combines it with the corresponding point feature through range-guided weighted aggregation:
\begin{equation}
\begin{aligned}
\mathbf{m}
&=
\operatorname{FtP}
\left(
\mathbf{F}_{stam},(u,v)
\right),\\
s_f
&=
\operatorname{Score}(d,\mathbf{m}),
\qquad
s_p
=
\operatorname{Score}(d,\mathbf{F}^{3D}),\\
[\alpha^{f},\alpha^{p}]
&=
\operatorname{Softmax}
\left(
[s_f,s_p]
\right),\\
\hat{\mathbf{F}}^{3D}
&=
\Phi_p
\left(
\alpha^{f}\mathbf{m}
+
\alpha^{p}\mathbf{F}^{3D}
\right).
\end{aligned}
\end{equation}

where $\operatorname{FtP}(\cdot)$ retrieves the frustum feature at the
projected location, $\operatorname{Score}(\cdot)$ computes the
range-conditioned matching score, and $\Phi_p(\cdot)$ denotes the
point-side fusion operation. The Softmax-normalized weights
$\alpha^{f}$ and $\alpha^{p}$ determine the relative contributions of the
frustum and point features, respectively, allowing contextual
information from $\mathbf{F}_{stam}$ to be integrated with the
geometric information encoded in $\mathbf{F}^{3D}$.

Point-to-Frustum interaction integrates the updated point-level geometry
with the contextual representation produced by STAM. Projection
correspondence organizes $\hat{\mathbf{F}}^{3D}$ in the frustum space,
while the corresponding range information is encoded as a geometric
feature. Joint fusion is formulated as
\begin{equation}
\hat{\mathbf{F}}_{stam}
=
\Phi_f\left(
\left[
\mathbf{F}_{stam},
\psi_d(range),
\operatorname{PtF}\left(\hat{\mathbf{F}}^{3D}\right)
\right]
\right),
\end{equation}

where $\operatorname{PtF}(\cdot)$ denotes the projection-based
organization of point features in the frustum space, $\psi_d(\cdot)$
encodes the range information, and $\Phi_f(\cdot)$ performs joint
convolutional fusion. The resulting
$\hat{\mathbf{F}}_{stam}$ combines frustum-level context,
point-level geometric structure, and range information within a unified
representation.

\subsection{Fusion Head Module}

Multi-stage point-frustum feature interaction progressively updates the point and frustum features, yielding hierarchical representations with varying spatial resolutions and receptive fields. These hierarchical representations encode spatial information at different granularities, with high-resolution features preserving finer spatial details and features from deeper stages capturing responses over broader receptive fields. The fusion head aggregates the multi-stage point and frustum features to integrate these complementary representations into unified point and frustum representations. Specifically, the point features are concatenated and fused according to point-wise correspondences, whereas the frustum features are first upsampled to a common spatial resolution and then concatenated and fused. The fused frustum features are subsequently mapped back to the 3D point domain and further integrated with the fused point features for final point-wise semantic prediction.

\section{Experiments}
In this section, we conduct systematic experiments to evaluate the effectiveness of RaSFNet for outdoor LiDAR point cloud semantic segmentation. Comparisons with state-of-the-art methods and ablation studies are conducted to assess its overall segmentation performance and the contributions of individual components, while qualitative results are provided to further demonstrate its segmentation performance.

\subsection{Datasets}
We conduct comprehensive evaluations on two widely used outdoor LiDAR semantic segmentation datasets, SemanticKITTI \cite{behley2019semantickitti} and nuScenes \cite{Panoptic-nuScenes}, covering diverse real-world autonomous driving scenarios. SemanticKITTI \cite{behley2019semantickitti} was collected in Karlsruhe, Germany, using a Velodyne HDL-64E LiDAR sensor, with a vertical field of view (FoV) ranging from $-25^\circ$ to $3^\circ$. Following the official split, 19,130 LiDAR scans are used for training and 4,071 for validation. Each scan contains approximately 120,000 points, with 19 semantic categories considered for evaluation. nuScenes was collected using a 32-beam LiDAR sensor with a vertical FoV ranging from $-30^\circ$ to $10^\circ$. It contains 28,130 training samples and 6,019 validation samples from 1,000 driving scenes. The dataset provides annotations across 32 semantic categories, of which 16 categories are evaluated for semantic segmentation.

\subsection{Evaluation Metrics}

We evaluate segmentation performance using the mean Intersection over Union (mIoU) and mean Accuracy (mAcc), which are commonly adopted for semantic segmentation. For each semantic class $c$, the IoU is defined as:
\begin{equation}
\mathrm{IoU}_c =
\frac{TP_c}
{TP_c + FP_c + FN_c},
\end{equation}
where $TP_c$, $FP_c$, and $FN_c$ denote the numbers of true-positive, false-positive, and false-negative points for class $c$, respectively. The mean IoU over $C$ evaluated classes is computed as:
\begin{equation}
\mathrm{mIoU} =
\frac{1}{C}
\sum_{c=1}^{C}
\mathrm{IoU}_c.
\end{equation}

The class-wise accuracy is defined as:
\begin{equation}
\mathrm{Acc}_c =
\frac{TP_c}
{TP_c + FN_c},
\end{equation}
and the mean Accuracy is given by:
\begin{equation}
\mathrm{mAcc} =
\frac{1}{C}
\sum_{c=1}^{C}
\mathrm{Acc}_c.
\end{equation}

\begin{table*}[!b]
\caption{The class-wise IoU scores of different LiDAR semantic segmentation approaches on the SemanticKITTI~\cite{behley2019semantickitti} val set. All mIoU scores are given in percentage (\%). The best and second best scores for each class are highlighted in bold and underline.}
\vspace{-0.1cm}
\begin{adjustbox}{width=\linewidth}
\begin{tabular}{rcccccccccccccccccccc}
\toprule
\textbf{Method} & {\textbf{mIoU}
} & {\textbf{car}} &{\textbf{bi.cle}} & {\textbf{mt.cle}} & {\textbf{truck}} & {\textbf{oth.v.}} & {\textbf{pers.}} & {\textbf{b.clst}} & {\textbf{m.clst}} & {\textbf{road}} & {\textbf{park.}} &{\textbf{sidew.}} & {\textbf{oth.g}} & {\textbf{build.}} & {\textbf{fence}} & {\textbf{veget.}} & {\textbf{trunk}} &{\textbf{terra.}} & {\textbf{pole}} & {\textbf{traff.}}
\\
\midrule
RandLA-Net~\cite{hu2020randla} & $50.0$ & $92.0$ & $8.0$ & $12.8$ & $74.8$ & $46.7$ & $52.3$ & $46.0$ & $0.0$ & $93.4$ & $32.7$ & $73.4$ & $0.1$ & $84.0$ & $43.5$ & $83.7$ & $57.3$ & $73.1$ & $48.0$ & $27.0$
\\
RangeNet++ \cite{milioto2019rangenet++} & $51.2$ & $89.4$ & $26.5$ & $48.4$ & $33.9$ & $26.7$ & $54.8$ & $69.4$ & $0.0$ & $92.9$ & $37.0$ & $69.9$ & $0.0$ & $83.4$ & $51.0$ & $83.3$ & $54.0$ & $68.1$ & $49.8$ & $34.0$
\\
SqueezeSegV2~\cite{wu2019squeezesegv2} & $40.8$ & $82.7$ & $15.1$ & $22.7$ & $25.6$ & $26.9$ & $22.9$ & $44.5$ & $0.0$ & $92.7$ & $39.7$ & $70.7$ & $0.1$ & $71.6$ & $37.0$ & $74.6$ & $35.8$ & $68.1$ & $21.8$ & $22.2$
\\
SqueezeSegV3~\cite{xu2020squeezesegv3} & $53.3$ & $87.1$ & $34.3$ & $48.6$ & $47.5$ & $47.1$ & $58.1$ & $53.8$ & $0.0$ & $95.3$ & $43.1$ & $78.2$ & $0.3$ & $78.9$ & $53.2$ & $82.3$ & $55.5$ & $70.4$ & $46.3$ & $33.2$
\\
SalsaNet~\cite{aksoy2020salsanet} & $59.4$ &$90.5$ & $44.6$ & $49.6$ & $86.3$ & $54.6$ & $74.0$ & $81.4$ & $0.0$ & $93.4$ & $40.6$ & $69.1$ & $0.0$ & $84.6$ & $53.0$ & $83.6$ & $64.3$ & $64.2$ & $54.4$ & $39.8$
\\
PMF~\cite{zhuang2021perception} & $63.9$ & $95.4$ & $47.8$ & $62.9$ & $68.4$ & $\mathbf{75.2}$ & $78.9$ & $71.6$ & $0.0$ & $\mathbf{96.4}$ & $43.5$ & $80.5$ & $1.0$ & $88.7$ & $60.1$ & $88.6$ & $\underline{72.7}$ & $75.3$ & $65.5$ & $43.0$
\\
RangeVit~\cite{ando2023rangevit} & $60.9$ & $94.7$ & $44.1$ & $61.4$ & $71.9$ & $37.7$ & $65.3$ & $75.5$ & $0.0$ & $95.5$ & $48.4$ & $83.1$ & $0.0$ & $88.3$ & $60.0$ & $86.3$ & $65.3$ & $72.7$ & $63.1$ & $42.7$
\\
MSF-CSCNet~\cite{MSF-CSCNet} & $65.8$ & $96.8$ & $50.9$ & $72.7$ & $90.4$ & $61.7$ & $76.2$ & $90.2$ & $0.0$ & $94.3$ & $44.8$ & $81.3$ & $0.5$ & $90.3$ & $59.2$ & $86.9$ & $66.5$ & $71.3$ & $65.5$ & $51.1$
\\
SphereFormer~\cite{lai2023spherical} & $67.8$ & $96.8$ & $50.7$ & $75.1$ & $\underline{93.1}$ & $65.7$ & $76.9$ & $92.0$ & $0.0$ & $94.7$ & $53.2$ & $82.2$ & $3.2$ & $90.7$ & $58.4$ & $\underline{88.7}$ & $71.2$ & $\underline{75.9}$ & $64.6$ & $\underline{54.5}$
\\
FRNet~\cite{xu2025frnet} & $67.6$ & $\mathbf{97.2}$ & $53.3$ & $72.9$ & $81.5$ & $\underline{72.9}$ & $77.2$ & $90.8$ & $\underline{0.2}$ & $95.9$ & $53.7$ & $\underline{83.9}$ & $9.0$ & $90.5$ & $65.9$ & $87.0$ & $66.8$ & $72.6$ & $64.0$ & $47.9$
\\
WaffleIron~\cite{puy2023using} & $68.0$ & $96.1$ & $58.1$ & $\underline{79.7}$ & $77.4$ & $59.0$ & $\underline{81.1}$ & $92.2$ & $\mathbf{1.3}$ & $95.5$ & $50.2$ & $83.6$ & $6.0$ & $\mathbf{92.1}$ & $\underline{67.5}$ & $87.8$ & $\mathbf{73.8}$ & $73.0$ & $65.7$ & $52.2$
\\
FARVNet~\cite{FARVNet} & $\underline{68.5}$ & $97.0$ & $54.2$ & $75.9$ & $89.6$ & $72.6$ & $76.0$ & $90.1$ & $0.0$ & $95.7$ & $\underline{56.9}$ & $83.4$ & $\mathbf{22.7}$ & $89.8$ & $62.3$ & $87.0$ & $65.9$ & $73.0$ & $63.2$ & $47.1$
\\
PC-BEV~\cite{pcbev}& $66.4$ & $ 94.3 $ & $\mathbf{67.6}$ & $71.4$ & $75.4$ & $67.5$ & $77.4$ & $81.3$ & $0.1$ & $95.2$ & $39.7$ & $83.0$ & $0.1$ & $86.3$ & $49.7$ & $85.1$ & $\mathbf{70.2}$ & $68.2$ & $\mathbf{70.2}$ & $\mathbf{64.2}$
\\
RangeRet~\cite{mosco2026revisiting} & $65.7$ & $93.6$ & $47.2$ & $65.4$ & $89.9$ & $69.1$ & $73.0$ & $83.7$ & $0.0$ & $94.8$ & $54.1$ & $81.3$ & $15.9$ & $87.3$ & $59.8$ & $86.0$ & $64.3$ & $72.8$ & $60.4$ & $50.1$
\\
MFINet~\cite{ma2026mfinet} & $67.7$ & $ 96.9 $ & $56.0$ & $79.0$ & $92.4$ & $68.0$ & $75.5$ & $\underline{94.4}$ & $0.0$ & $93.6$ & $49.6$ & $81.3$ & $0.0$ & $91.0$ & $63.8$ & $\mathbf{89.1}$ & $64.3$ & $\mathbf{77.2}$ & $64.7$ & $49.6$
\\
VaViT ~\cite{VaViT} & $67.6$ & $95.7$ & $56.5$ & $74.9$ & $\mathbf{95.4}$ & $52.1$ & $\mathbf{81.4}$ & $\mathbf{94.5}$ & $0.1$ & $95.6$ & $51.1$ & $83.8$ & $0.1$ & $90.6$ & $61.0$ & $88.1$ & $\underline{72.0}$ & $73.5$ & $\underline{65.9}$ & $51.6$ 
\\
RaSFNet & $\mathbf{70.2}$ & $ \underline{97.1} $ & $\underline{58.9}$ & $\mathbf{80.8}$ & $84.6$ & $70.7$ & $81.0$ & $92.8$ & $0.0$ & $\underline{96.0}$ & $\mathbf{57.6}$ & $\mathbf{85.4}$ & $\underline{20.1}$ & $\underline{92.0}$ & $\mathbf{71.1}$ & $87.9$ & $68.6$ & $74.2$ & $65.5$ & $49.4$

\\\bottomrule
\end{tabular}
\end{adjustbox}
\label{table:semantickitti-class}
\end{table*}

\subsection{Implementation Details}

Although SemanticKITTI \cite{behley2019semantickitti} and nuScenes \cite{Panoptic-nuScenes} differ in LiDAR beam count (64 vs. 32), both utilize a full $360^{\circ}$ horizontal FoV. The range image resolutions were set to $64\times1024$ and $32\times1024$ for SemanticKITTI \cite{behley2019semantickitti} and nuScenes \cite{Panoptic-nuScenes}, respectively. Each projected point is represented by its 3D coordinates and reflectance intensity. The feature encoder uses channel dimensions of (64,128,256,256) and compresses the output to 20 channels, while the four backbone stages each output 128 channels. The model is trained for 250,000 iterations using AdamW with an initial learning rate of 0.001, a batch size of 4, and the OneCycle learning-rate policy. All experiments were conducted on two NVIDIA RTX 4090 GPUs.

\subsection{Quantitative Results}

We compare RaSFNet with LiDAR semantic segmentation methods based on different representations, including point-based, sparse-voxel, bird\textquotesingle s-eye-view, and range-view representations, on the SemanticKITTI \cite{behley2019semantickitti} and nuScenes \cite{Panoptic-nuScenes} validation sets. The results are presented in Tables~\ref{table:semantickitti-class} and~\ref{table:nuscenes-class-val}, respectively. RaSFNet achieves mIoU scores of 70.2\% and 77.7\% on the SemanticKITTI \cite{behley2019semantickitti} and nuScenes \cite{Panoptic-nuScenes} validation sets, respectively. On SemanticKITTI \cite{behley2019semantickitti}, RaSFNet performs well in the motorcycle, parking, sidewalk, and fence classes. On nuScenes \cite{Panoptic-nuScenes}, RaSFNet also obtains favorable results in the bicycle, motorcycle, bus, and pedestrian classes. The results suggest that RaSFNet maintains competitive overall accuracy and performs well across several semantic classes.

\begin{table*}[!htbp]
\caption{Class-wise IoU (\%) on the nuScenes validation set~\cite{Panoptic-nuScenes}. Best and second-best distinct scores in each column are shown in bold and underlined, respectively; ties share the same rank.}
\label{table:nuscenes-class-val}
\vspace{-0.1cm}
\begin{adjustbox}{width=\linewidth}
\begin{tabular}{rccccccccccccccccc}
\toprule
\textbf{Method} & \textbf{mIoU} & \textbf{bar.} & \textbf{bi.cle} & \textbf{bus} & \textbf{car} & \textbf{con.v.} & \textbf{mt.cle} & \textbf{ped.} & \textbf{t.cone} & \textbf{trail.} & \textbf{truck} & \textbf{dr.s.} & \textbf{oth.f.} & \textbf{sidew.} & \textbf{terra.} & \textbf{man.} & \textbf{veget.} \\
\midrule
RangeNet++~\cite{milioto2019rangenet++} & $65.5$ & $66.0$ & $21.3$ & $77.2$ & $80.9$ & $30.2$ & $66.8$ & $69.6$ & $52.1$ & $54.2$ & $72.3$ & $94.1$ & $66.6$ & $63.5$ & $70.1$ & $83.1$ & $79.8$ \\
PolarNet~\cite{zhang2020polarnet} & $71.5$ & $74.7$ & $28.2$ & $85.3$ & $90.9$ & $35.1$ & $77.5$ & $71.3$ & $58.8$ & $57.4$ & $76.1$ & $96.5$ & $71.1$ & $74.7$ & $74.0$ & $87.3$ & $85.7$ \\
PCSCNet~\cite{park2023pcscnet} & $72.0$ & $73.3$ & $42.2$ & $87.8$ & $86.1$ & $44.9$ & $\underline{82.2}$ & $76.1$ & $62.9$ & $49.3$ & $77.3$ & $95.2$ & $66.9$ & $69.5$ & $72.3$ & $83.7$ & $82.5$ \\
MinkowskiNet~\cite{choy20194d} & $73.5$ & $74.5$ & $39.8$ & $89.7$ & $\underline{91.0}$ & $44.3$ & $78.7$ & $77.3$ & $60.4$ & $56.7$ & $82.6$ & $95.5$ & $68.8$ & $71.3$ & $74.4$ & $86.1$ & $84.9$ \\
SalsaNext~\cite{SalsaNext} & $72.0$ & $74.8$ & $34.1$ & $85.9$ & $88.4$ & $42.2$ & $72.4$ & $72.2$ & $63.1$ & $61.3$ & $76.5$ & $96.0$ & $70.8$ & $71.2$ & $71.5$ & $86.7$ & $84.4$ \\
CENet~\cite{cheng2022cenet} & $73.3$ & $75.8$ & $36.4$ & $87.7$ & $88.5$ & $44.6$ & $73.4$ & $75.2$ & $65.1$ & $62.3$ & $77.5$ & $96.2$ & $71.8$ & $72.3$ & $74.5$ & $86.8$ & $85.4$ \\
SVASeg~\cite{zhao2022svaseg} & $74.7$ & $73.1$ & $44.5$ & $88.4$ & $86.6$ & $48.2$ & $80.5$ & $\underline{77.7}$ & $65.6$ & $57.5$ & $82.1$ & $96.5$ & $70.5$ & $74.7$ & $74.6$ & $87.3$ & $86.9$ \\
RangeViT~\cite{ando2023rangevit} & $75.2$ & $75.5$ & $40.7$ & $88.3$ & $90.1$ & $49.3$ & $79.3$ & $77.2$ & $66.3$ & $65.2$ & $80.0$ & $96.4$ & $71.4$ & $73.8$ & $73.8$ & $\underline{89.9}$ & $87.2$ \\
MSF-CSCNet~\cite{MSF-CSCNet} & $75.2$ & $75.1$ & $40.3$ & $91.3$ & $88.3$ & $45.8$ & $78.7$ & $\mathbf{78.7}$ & $64.5$ & $64.1$ & $80.9$ & $96.5$ & $73.2$ & $75.0$ & $75.6$ & $88.1$ & $87.3$ \\
FRNet~\cite{xu2025frnet} & $76.1$ & $77.2$ & $39.5$ & $93.4$ & $88.6$ & $52.1$ & $81.4$ & $75.1$ & $65.7$ & $66.2$ & $79.7$ & $96.8$ & $\underline{75.3}$ & $75.4$ & $\underline{75.9}$ & $88.4$ & $85.4$ \\
WaffleIron~\cite{puy2023using} & $76.1$ & $\mathbf{77.8}$ & $\underline{45.8}$ & $\underline{93.7}$ & $85.3$ & $50.2$ & $81.9$ & $75.8$ & $66.2$ & $\mathbf{68.7}$ & $\underline{83.2}$ & $96.7$ & $72.9$ & $74.9$ & $73.7$ & $86.3$ & $84.0$ \\
HARP-NeXt~\cite{abouhaidar2025harpnext} & $\underline{77.1}$ & $76.2$ & $37.5$ & $\mathbf{94.2}$ & $\mathbf{91.1}$ & $\mathbf{60.8}$ & $59.5$ & $75.9$ & $\mathbf{73.7}$ & $65.0$ & $\mathbf{84.8}$ & $\mathbf{97.7}$ & $\mathbf{75.9}$ & $\mathbf{79.4}$ & $\mathbf{83.3}$ & $\mathbf{91.0}$ & $\mathbf{88.0}$ \\
FLARES (FIDNet)~\cite{yang2026flares} & $76.6$ & $77.6$ & $43.5$ & $92.9$ & $88.1$ & $56.5$ & $79.5$ & $\underline{77.7}$ & $65.3$ & $67.0$ & $83.1$ & $96.6$ & $72.8$ & $75.0$ & $74.5$ & $88.5$ & $86.8$ \\
RaSFNet & $\mathbf{77.7}$ & $\underline{77.7}$ & $\mathbf{46.2}$ & $\mathbf{94.2}$ & $87.9$ & $\underline{58.6}$ & $\mathbf{87.2}$ & $\mathbf{78.7}$ & $\underline{68.1}$ & $\underline{68.1}$ & $78.2$ & $\underline{97.0}$ & $73.6$ & $\underline{75.6}$ & $74.3$ & $89.5$ & $\underline{87.5}$ \\
\bottomrule
\end{tabular}
\end{adjustbox}
\end{table*}

As the sensing distance increases, the spatial sampling density of LiDAR point clouds gradually decreases, accompanied by an overall decline in semantic segmentation performance across different methods. Near-range regions generally yield higher mIoU than far-range regions, indicating the performance variation under different point-density conditions. Table~\ref{tab:range-density} provides a quantitative comparison of semantic segmentation performance and corresponding point density across different sensing-distance ranges on the SemanticKITTI \cite{behley2019semantickitti} validation set. The average point density decreases from 923.8 to 13.1 points/m$^3$ with increasing sensing distance. Our method achieves mIoUs of 77.5\%, 71.3\%, 65.9\%, 59.9\%, and 45.3\% in the 0-5 m, 5-10 m, 10-20 m, 20-40 m, and 40-50 m ranges, respectively, achieving the highest mIoU among the compared methods in all five ranges. For the 50-60 m range, our method achieves an mIoU of 25.4\%, close to the highest value of 26.0\%. Overall, favorable segmentation performance is maintained in both high-density near-range regions and low-density middle- and far-range regions. Distance-aware modeling of image feature extraction and point-image feature interaction facilitates adaptation to feature representation variations across different sensing distances.

\begin{table}[!htbp]
\centering
\scriptsize
\caption{Distance-wise semantic segmentation performance (mIoU, \%) on the SemanticKITTI \cite{behley2019semantickitti} validation set under different point-density levels. Point density is computed as the average number of points per non-empty 0.5 m$^3$ voxel within each distance range across all frames.}
\label{tab:range-density}
\setlength{\tabcolsep}{3pt}
\resizebox{\linewidth}{!}{
\begin{tabular}{l c c c c c c} 
\toprule
\textbf{Method} & \textbf{0-5 m} & \textbf{5-10 m} & \textbf{10-20 m} &\textbf{20-40 m} & \textbf{40-50 m} & \textbf{50-60 m} \\
\textbf{Density} & \textbf{923.8} & \textbf{331.9} & \textbf{88.2} & \textbf{29.8} & \textbf{16.2} & \textbf{13.1} \\
\midrule
SqueezeSegV2~\cite{wu2019squeezesegv2} & 43.6 & 43.1 & 38.3 & 27.7 & 19.5 & 9.3\\
FIDNet~\cite{zhao2021fidnet} & 56.0 & 61.3 &58.5 &48.4 &36.0 &21.1\\
RangeVit~\cite{ando2023rangevit} &58.6 & 62.7 &60.6 & 50.5 & 40.7 & 23.3\\
SegNet4D~\cite{wang2025segnet4d} & 65.4 & 62.8 &61.7 &56.0 &43.3 &26.0\\
Cylinder3D~\cite{zhu2021cylindrical} & 68.4 & 66.6 & 63.8 &58.7 & 42.2 &15.2\\
WaffleIron~\cite{puy2023using} & 72.8 & 68.7 & 66.1 & 60.3 & 41.7 & 11.6\\
FRNet~\cite{xu2025frnet} & 73.9 & 68.7 &64.8 & 55.6 & 41.5 & 24.2\\
FARVNet~\cite{FARVNet} & 73.7 & 68.4 &65.6 & 57.7 & 40.3 & 22.6\\
RangeRet~\cite{mosco2026revisiting} & 75.0 & 67.9 &62.1 &48.8 &35.7 & 14.2 \\
Ours &77.5 & 71.3 & 65.9 & 59.9 & 45.3 & 25.4\\
\bottomrule
\end{tabular}
}
\end{table}

\subsection{Qualitative Results}
\begin{figure}[!h] \centering \begin{adjustbox}{max width=\linewidth} \begin{tikzpicture} \node[inner sep=0pt, anchor=south west] (img) {\includegraphics[width=\linewidth]{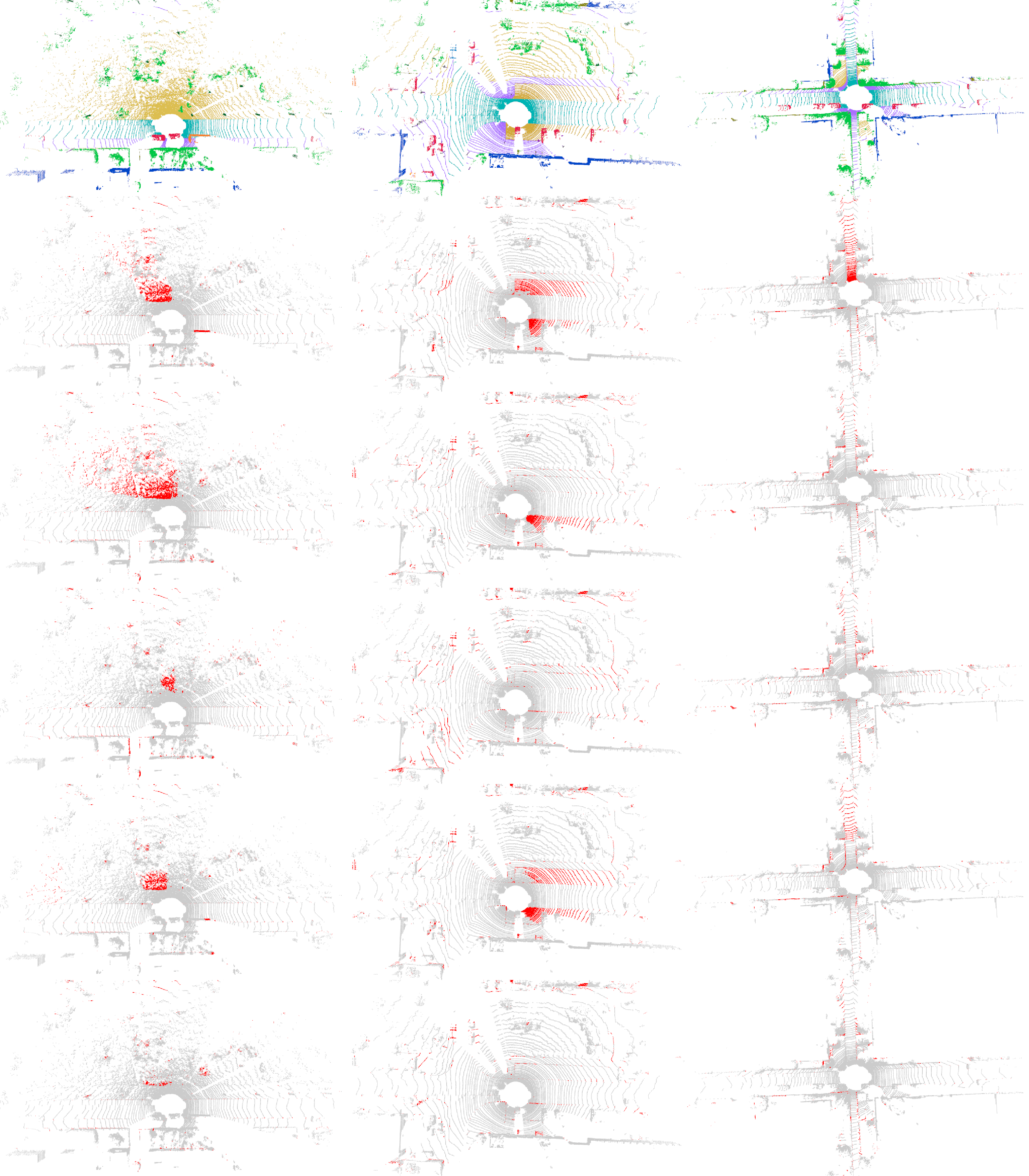}}; \begin{scope}[x={(img.south east)}, y={(img.north west)}] \def\nrows{6} \def\xlabel{-0.05} \def\yoffset{-10/\nrows} \foreach \i/\name/\color in { 1/\textbf{GT}/cyan!10, 2/\textbf{SphereFormer}/blue!15, 3/\textbf{FRNet}/green!15, 4/\textbf{RangeRet}/orange!15, 
5/\textbf{VaViT}/orange!15,
6/\textbf{Ours}/purple!15 }{ \pgfmathsetmacro{\ypos}{1 - (\i - 0.5)/\nrows} \node[ rectangle, rounded corners=2pt, fill=\color, minimum width=1.3cm, minimum height={\dimexpr1cm/\nrows}, inner sep=1.5pt, align=center, anchor=center, rotate=90, transform shape, draw=none, font=\small ] at (\xlabel,\ypos) {\name}; } \def\ncols{3} \foreach \j/\label in { 1/{(a)}, 2/{(b)}, 3/{(c)}}{ \pgfmathsetmacro{\xpos}{(\j - 0.5)/\ncols} \node[anchor=north] at (\xpos, -0.00) {\label}; } \end{scope} \end{tikzpicture} \end{adjustbox} \caption{ Segmentation errors in challenging scenes on the SemanticKITTI \cite{behley2019semantickitti} val set. Red represents misclassified areas, and gray represents correctly classified areas.} \label{qualitative_results_error} \end{figure}

Maintaining spatial consistency remains challenging in LiDAR semantic segmentation, particularly within large-scale semantic regions. Existing methods may generate locally clustered errors that spread across semantic boundaries, resulting in fragmented predictions. Fig.~\ref{qualitative_results_error} presents qualitative comparisons in three representative scenarios: a densely vegetated area, a road-terrain boundary, and a crossroad intersection. Across these scenarios, the compared methods exhibit clustered errors within vegetation or continuous misclassifications along road boundaries and connecting branches. RaSFNet produces fewer misclassified points, with more scattered errors and more coherent predicted regions. Its design considers both local geometric differences and global spatial relationships, allowing semantic discrimination to draw on fine-grained structural details and broader contextual information.

\begin{figure}[!t]
    \centering
    \begin{adjustbox}{max width=\linewidth}
        \begin{tikzpicture}
            \node[inner sep=0pt, anchor=south west] (img)
            {\includegraphics[width=\linewidth]{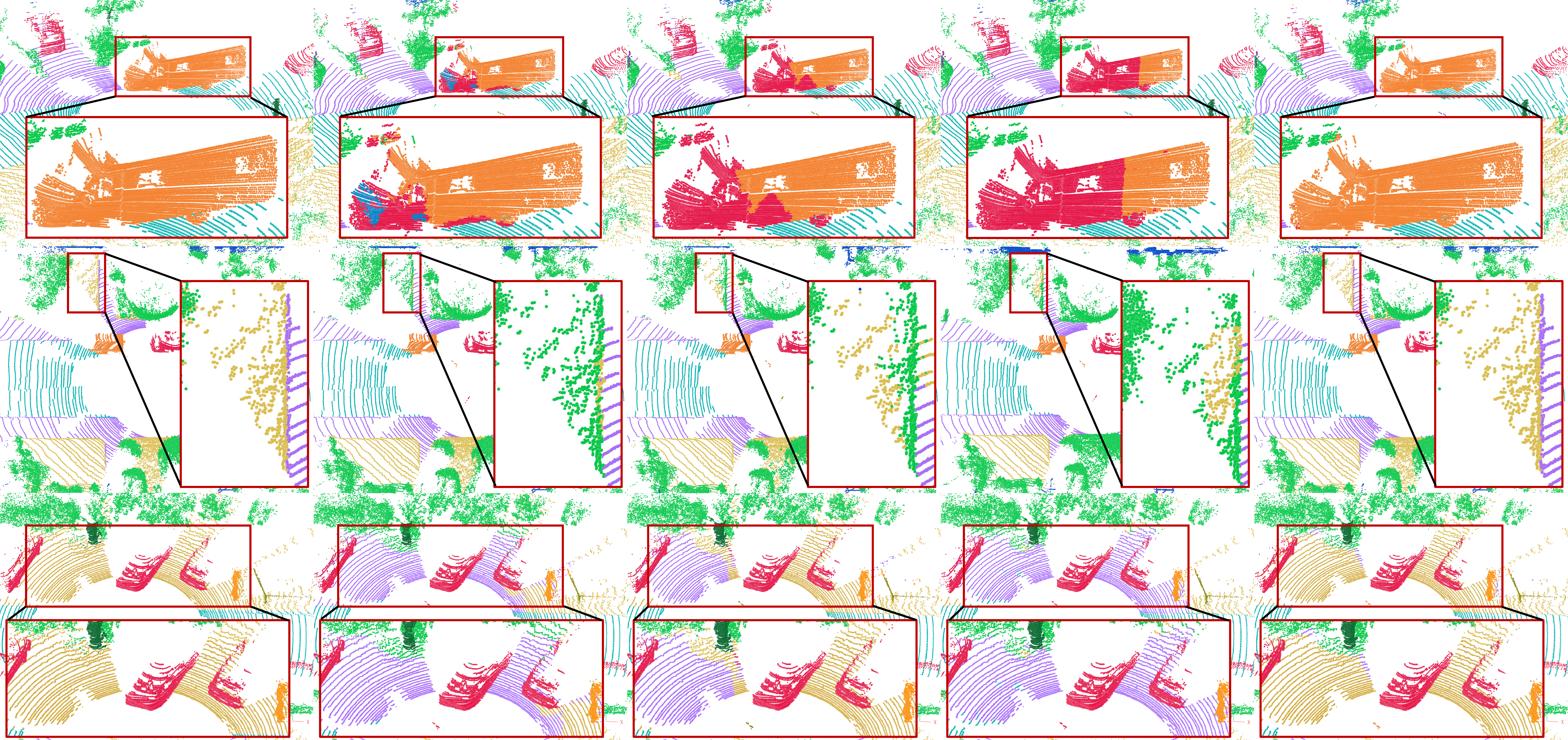}};
            \begin{scope}[
                x={(img.south east)},
                y={(img.north west)}
            ]
            \def\ncols{5}
            \foreach \j/\name/\color in {
                1/\textbf{GT}/cyan!10,
                2/\textbf{FRNet}/green!15,
                3/\textbf{RangeRet}/orange!15,
                4/\textbf{VaViT}/orange!15,
                5/\textbf{Ours}/purple!15
            }{
                \pgfmathsetmacro{\xpos}{(\j - 0.5)/\ncols}
                \node[
                rectangle,
                rounded corners=2pt,
                fill=\color,
                inner sep=2pt,
                text height=1.5ex,
                text depth=0.3ex,
                align=center,
                anchor=south,
                draw=none,
                font=\small
            ] at (\xpos,1.01) {\name};
            }
            \def\nrows{3}
            \foreach \i/\label in {
                1/{(a)},
                2/{(b)},
                3/{(c)}
            }{
                \pgfmathsetmacro{\ypos}{1 - (\i - 0.5)/\nrows}
                \node[
                    anchor=east,
                    font=\small
                ] at (-0.01,\ypos) {\label};
            }
            \end{scope}
        \end{tikzpicture}
    \end{adjustbox}
    \caption{
        Qualitative comparison of semantic segmentation results on the SemanticKITTI~\cite{behley2019semantickitti} validation set.
    }
    \label{qualitative_results}
\end{figure}

To evaluate the ability of RaSFNet to distinguish semantically different classes with similar local geometric structures, Fig.~\ref{qualitative_results} presents qualitative comparisons with FRNet~\cite{xu2025frnet}, RangeRet~\cite{mosco2026revisiting}, and VaViT~\cite{VaViT} in three challenging scenes. Incomplete observations and occlusion obscure distinctive geometric characteristics, while similar planar geometry makes terrain and sidewalk difficult to distinguish. Across these scenes, the compared methods exhibit confusion between bus and car, terrain and vegetation, or terrain and sidewalk, whereas RaSFNet more accurately identifies the highlighted regions. RaSFNet enhances subtle geometric differences through encoding and guides the collaborative updating of 3D structural features and 2D frustum features under persistent spatial constraints, thereby improving semantic discrimination between regions with similar local geometric structures.

\begin{figure}[!h]
    \centering
    \begin{adjustbox}{max width=\linewidth}
        \begin{tikzpicture}
            \node[inner sep=0pt, anchor=south west] (img)
            {\includegraphics[width=\linewidth]
            {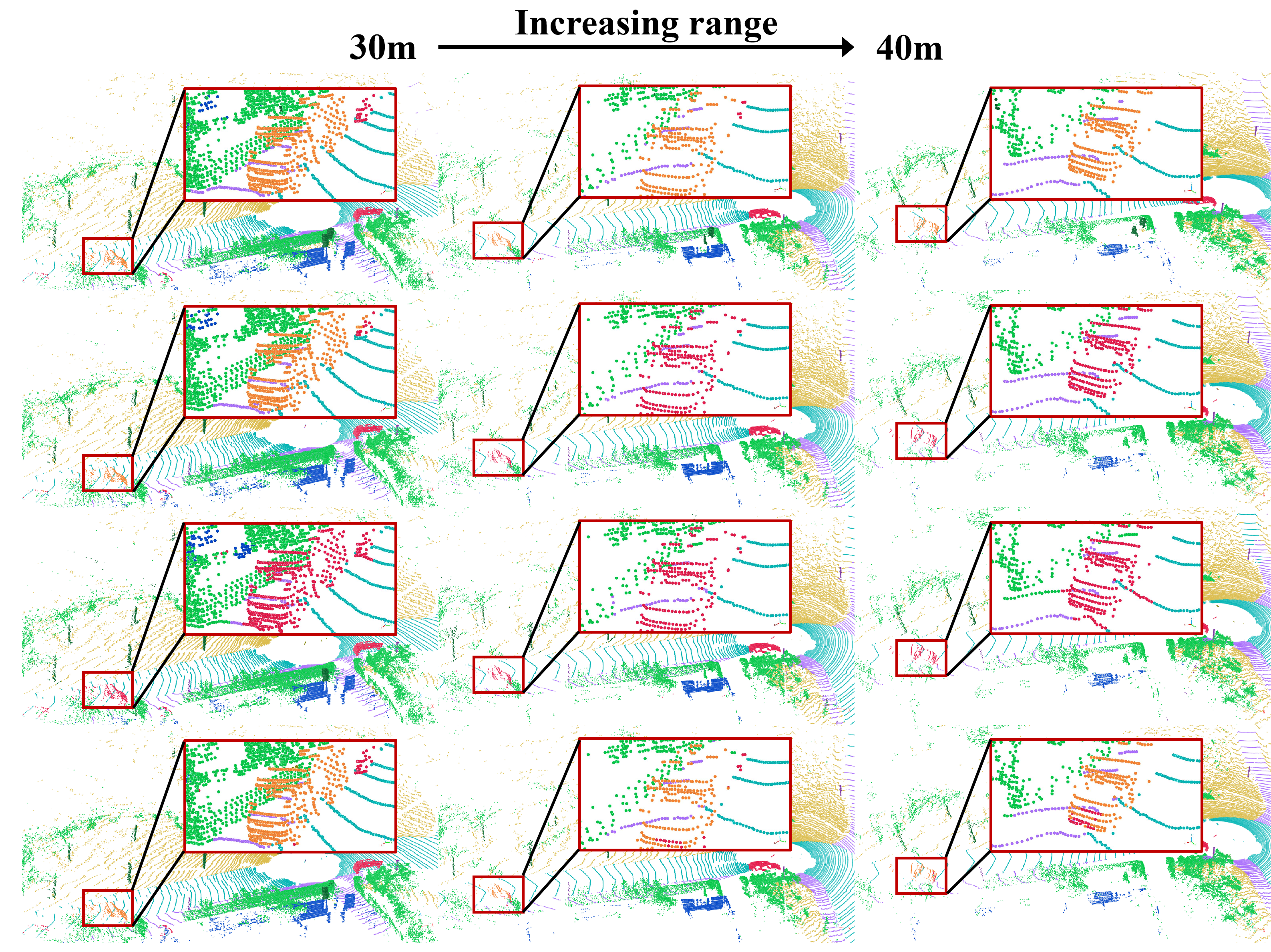}};

            \begin{scope}[
                x={(img.south east)},
                y={(img.north west)}
            ]
                \def\nrows{4}
                \def\xlabel{-0.025}
                \def\ylabelshift{0.05}
                \foreach \i/\name/\color in {
                    1/\textbf{GT}/cyan!10,
                    2/\textbf{FRNet}/green!15,
                    3/\textbf{RangeRet}/orange!15,
                    4/\textbf{Ours}/purple!15
                }{
                    \pgfmathsetmacro{\ypos}{
                        1-(\i-0.5)/\nrows-\ylabelshift
                    }
                    \node[
                        rectangle,
                        rounded corners=2pt,
                        fill=\color,
                        inner sep=2pt,
                        align=center,
                        anchor=center,
                        rotate=90,
                        transform shape,
                        draw=none,
                        font=\small
                    ] at (\xlabel,\ypos) {\name};
                }
            \end{scope}
        \end{tikzpicture}
    \end{adjustbox}
    \caption{
        Qualitative comparison of semantic segmentation results across a temporal sequence on the SemanticKITTI~\cite{behley2019semantickitti} validation set.
    }
    \label{qualitative_results_range}
\end{figure}

Long-range semantic perception remains challenging because increasing sensing distance substantially reduces the density of LiDAR measurements. Sparse measurements provide incomplete geometric structures and limited class-discriminative cues, increasing the likelihood of recognition errors. As shown in Fig.~\ref{qualitative_results_range}, segmentation results are compared for the same bus at progressively greater distances across consecutive frames. FRNet \cite{xu2025frnet} and RangeRet \cite{mosco2026revisiting} produce incorrect labels within the bus region, with segmentation completeness deteriorating as the sensing distance increases. By contrast, RaSFNet preserves a more complete bus region across the examined distances, demonstrating greater robustness to the feature degradation associated with sparse long-range observations.

\subsection{Ablation Study}

\textbf{Component Ablation:} Table~\ref{tab:ablation} evaluates the contributions of RAGDE, STAM, and RCPFF through leave-one-component-out experiments under identical training settings. Removing RCPFF reduces mIoU by 2.7 percentage points on SemanticKITTI and 2.2 percentage points on nuScenes, highlighting the importance of effectively integrating 3D structural information with 2D frustum features. Removing STAM also degrades performance on both datasets, suggesting that axis-aware aggregation and serpentine sequence modeling help capture contextual information from the encoded geometric features. Removing RAGDE reduces both mIoU and mAcc, supporting the contribution of ray-aligned encoding to local geometric representation. These consistent trends across the two datasets support the contributions of RAGDE, STAM, and RCPFF within the overall framework. The complete configuration outperforms all ablated variants in both mIoU and mAcc on both datasets.

\begin{table}[!h]
\centering
\caption{Ablation study of RAGDE, STAM, and RCPFF on the SemanticKITTI and nuScenes validation sets. All scores are reported in percentage (\%). w/o denotes the removal of the corresponding module.}
\label{tab:ablation}
\begin{tabular}{lcccc}
\toprule
\multirow{2}{*}{\textbf{Variant}}
& \multicolumn{2}{c}{\textbf{SemanticKITTI}}
& \multicolumn{2}{c}{\textbf{nuScenes}} \\
\cmidrule(lr){2-3}\cmidrule(lr){4-5}
& \textbf{mIoU} & \textbf{mAcc}
& \textbf{mIoU} & \textbf{mAcc} \\
\midrule
w/o RAGDE & 69.2 & 75.9 & 77.3 &84.4 \\
w/o STAM  & 68.5 & 74.7 & 76.7 & 84.3 \\
w/o RCPFF & 67.5 & 74.3 & 75.5 & 82.8 \\
RaSFNet    & 70.2 & 76.6 & 77.7 & 85.2 \\
\bottomrule
\end{tabular}
\end{table}

\textbf{Component Designs:} Table~\ref{tab:module-replacement} compares segmentation performance and parameter counts when the original modules are replaced with alternative encoding, sequence modeling, and fusion methods. The evaluated replacement configurations contain 8.1M-21.7M parameters and achieve mIoU scores of 66.1\%-68.6\%. In comparison, RaSFNet achieves 70.2\% mIoU and 76.6\% mAcc with 8.0M parameters, outperforming the evaluated alternatives. These differences are associated with how geometric information is modeled: the respective replacement methods do not explicitly characterize differences in range distributions and geometric organization along sensing rays, organize feature sequences according to LiDAR azimuthal sampling patterns, or maintain a shared range constraint throughout point-frustum feature updates. RaSFNet explicitly accounts for these geometric and sampling characteristics, enabling more discriminative feature encoding and spatially constrained point-frustum interaction.

\begin{table}[!h]
\centering
\caption{Module replacement study on the SemanticKITTI validation set. Each variant replaces only the specified module while retaining the remaining components. All scores are reported as percentages (\%).}
\label{tab:module-replacement}
\begin{tabular}{llccc}
\toprule
\textbf{Module} & \textbf{Replacement}
& \textbf{mIoU} & \textbf{mAcc} & \textbf{Params} \\
\midrule
\multirow{3}{*}{RAGDE}
& PointNet++ \cite{qi2017pointnet++} & 68.2 & 74.3 & 8.2M  \\
& Cylinder3D \cite{zhu2021cylindrical} & 68.6 & 74.7 & 8.2M \\
& MDSLCA \cite{guang2026mdslca} & 67.8 & 74.6 & 8.1M \\
\midrule
\multirow{2}{*}{STAM}
& Mamba \cite{gu2024mamba} & 67.9 & 74.0 & 10.0M \\
& Vision Mamba \cite{VisionMamba}  & 67.1 & 73.3 & 12.4M  \\
\midrule
\multirow{2}{*}{RCPFF}
& Concat-MLP Fusion & 66.1 & 72.6 & 21.7M  \\
& HARP-NeXt \cite{abouhaidar2025harpnext} & 68.5 & 74.5  & 8.4M \\
\midrule
\multicolumn{2}{l}{RaSFNet (original modules)}
& 70.2 & 76.6 & 8.0M \\
\bottomrule
\end{tabular}
\end{table}

\textbf{Range-Image Resolution:} Range-image resolution determines the angular coverage of each frustum and the distribution of projected points. Table~\ref{tab:resolution-ablation} compares different resolutions on both datasets. A low resolution assigns more points to each frustum, which may increase projection conflicts and obscure local structures near semantic boundaries. Increasing the resolution separates projected points more effectively, whereas an excessively fine grid creates sparsely occupied frustums and limits feature aggregation. Consequently, neither mIoU nor mAcc improves monotonically with increasing image width. Among the evaluated configurations, SemanticKITTI achieves the highest mIoU and mAcc at $64\times1024$, while nuScenes achieves the highest values at $32\times480$.

\begin{table}[!h]
\centering
\caption{Effect of range-image resolution on the SemanticKITTI and nuScenes validation sets. Resolution is expressed as $H \times W$. All scores are reported as percentages (\%).}
\label{tab:resolution-ablation}
\setlength{\tabcolsep}{4pt}
\begin{tabular}{cccccc}
\toprule
\multicolumn{3}{c}{\textbf{SemanticKITTI}}
& \multicolumn{3}{c}{\textbf{nuScenes}} \\
\cmidrule(lr){1-3}\cmidrule(lr){4-6}
\textbf{Resolution} & \textbf{mIoU} & \textbf{mAcc}
& \textbf{Resolution} & \textbf{mIoU} & \textbf{mAcc} \\
\midrule
$64 \times 256$  & 65.2 & 72.3 & $32 \times 256$  & 75.8 & 83.3 \\
$64 \times 512$ & 68.9 & 75.4 & $32 \times 480$  & 77.7 & 85.2\\
$64 \times 768$ & 68.3 & 74.8 & $32 \times 512$ & 77.3 & 84.4 \\
$64 \times 1024$ & 70.2 & 76.6 & $32 \times 768$ & 77.5 & 84.7 \\
$64 \times 1280$ & 68.7 & 75.8 & $32 \times 1024$ & 77.3 & 84.5 \\
\bottomrule
\end{tabular}
\end{table}

\section{Conclusion}
This paper presented RaSFNet for projection-based LiDAR semantic segmentation. Guided by the physical acquisition characteristics of LiDAR, RaSFNet amplifies subtle differences among similar local point distributions through ray-aligned encoding, serializes frustum features in a serpentine row-wise order along the azimuthal dimension, and uses range information shared by point and frustum representations to constrain their bidirectional updates. Across the evaluated datasets, RaSFNet produces more coherent semantic regions and clearer class boundaries, with less confusion between spatially adjacent classes that share similar geometric structures. Class-wise evaluations and ablation studies provide further analyses of the segmentation performance and the roles of the main components. Future work will explore extending RaSFNet to other 3D perception tasks, including 3D object detection and semantic occupancy prediction.
\bibliographystyle{IEEEtran}
\bibliography{sample}

\newpage

\end{document}